\documentclass[10pt,twocolumn,letterpaper]{article}

\usepackage{cvpr}
\usepackage{times}
\usepackage{epsfig}
\usepackage{graphicx}
\usepackage{amsmath}
\usepackage{amssymb}
\usepackage{multicol}
\usepackage{multirow}
\usepackage{enumitem}
\usepackage{nccmath}

\usepackage{booktabs}

\usepackage[breaklinks=true,bookmarks=false,linkcolor=red,colorlinks=true,citecolor=blue]{hyperref}

\cvprfinalcopy % *** Uncomment this line for the final submission

\def\cvprPaperID{****} % *** Enter the CVPR Paper ID here

\begin{document}

%%%%%%%%% TITLE
\title{DSGN: Deep Stereo Geometry Network for 3D Object Detection}

\author{Yilun Chen$^{1}$ \quad Shu Liu$^{2}$ \quad Xiaoyong Shen$^{2}$ \quad Jiaya Jia$^{1,2}$ \\ 
$^{1}$The Chinese University of Hong Kong\quad\quad$^{2}$SmartMore\\
{\tt\small \{ylchen, leojia\}@cse.cuhk.edu.hk} \quad\quad{\tt\small \{sliu, xiaoyong\}@smartmore.com}
}

\maketitle
% \thispagestyle{empty}

%%%%%%%%% ABSTRACT
\begin{abstract}
% \ylchen{check}
Most state-of-the-art 3D object detectors heavily rely on LiDAR sensors because there is a large performance gap between image-based and LiDAR-based methods. It is caused by the way to form representation for the prediction in 3D scenarios. Our method, called Deep Stereo Geometry Network (DSGN), significantly reduces this gap by detecting 3D objects on a differentiable volumetric representation -- 3D geometric volume, which effectively encodes 3D geometric structure for 3D regular space. With this representation, we learn depth information and semantic cues simultaneously. For the first time, we provide a simple and effective one-stage stereo-based 3D detection pipeline that jointly estimates the depth and detects 3D objects in an end-to-end learning manner. Our approach outperforms previous stereo-based 3D detectors (about 10 higher in terms of AP) and even achieves comparable performance with several LiDAR-based methods on the KITTI 3D object detection leaderboard. Our code is publicly available at \url{https://github.com/chenyilun95/DSGN}. 
\end{abstract}

%%%%%%%%% BODY TEXT

\section{Introduction}

3D scene understanding is a challenging task in 3D perception, which serves as a basic component for autonomous driving and robotics. Due to the great capability of LiDAR sensors to accurately retrieve 3D information, we witness fast progress on 3D object detection. Various 3D object detectors were proposed \cite{MV3D, AVOD, VoxelNet, MMF, contfuse, fpointnet, pointrcnn, std, fastpointrcnn} to exploit LiDAR point cloud representation. The limitation of LiDAR is on the relatively sparse resolution of data with several laser beams and on the high price of the devices. 

\begin{figure}
    \begin{center}
    \includegraphics[width=90mm]{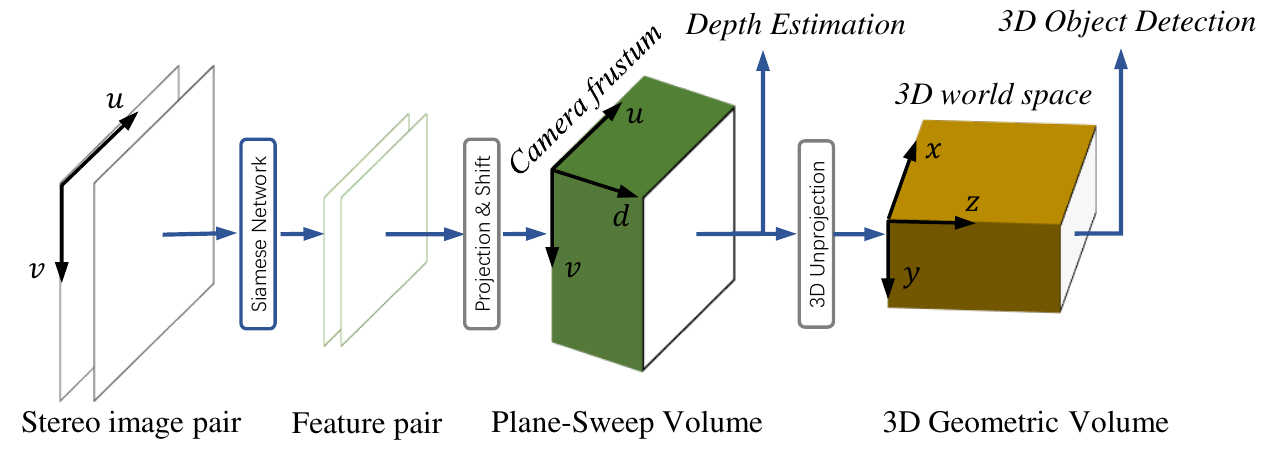}
    \end{center}\vspace{-0.1in}
    \caption{DSGN jointly estimates depth and detects 3D objects from a stereo image pair. It intermediately generates a plane-sweep volume and 3D geometric volume to represent 3D structure in two different 3D space. }
    \label{fig:small_pipeline}
\end{figure}

In comparison, video cameras are cheaper and are with much denser resolutions. The way to compute scene depth on stereo images is to consider disparity via stereo correspondence estimation. Albeit recently several 3D detectors based on either monocular \cite{qin2019monogrnet, 3dop, monocular3d, accuratemono3d, MLF} or stereo \cite{stereorcnn, pseudolidar, triangulation, pseudo++} setting push the limit of image-based 3D object detection, the accuracy is still left far behind compared with the LiDAR-based approaches. 

%The perception range depends on the setting of stereo cameras, \ie., image resolution, the focal length and baseline of camera. In this way, we believe that stereo-based vision method has the potential to provide another low-cost method or compensate for autonomous driving. 

\vspace{0.1in}
\noindent{\bf Challenges~~} One of the greatest challenges for image-based approaches is to give appropriate and effective representation for predicting 3D objects. Most recent work  \cite{stereorcnn, qin2019monogrnet,MLF,triangulation,disentangling,brazil2019m3d} divides this task into two sub ones, \ie, depth prediction and object detection. Camera projection is a process that maps 3D world into a 2D image. One 3D feature in different object poses causes local appearance changes, making it hard for a 2D network to extract stable 3D information. 
%Besides, these approaches require special design to combine two branches of information. 

Another line of solutions \cite{pseudolidar, pseudo++, monopseudoliar, accuratemono3d} generate intermediate point cloud followed by a LiDAR-based 3D object detector. This 3D representation is less effective since the transformation is non-differentiable and incorporates several independent networks. Besides, the point cloud faces the challenge of object artifacts \cite{depthcoefficients, monopseudoliar, pseudo++} that limits the detection accuracy of the following 3D object detector.

\vspace{0.1in}
\noindent{\bf Our Solution~~}  
% advantage of representation
In this paper, we propose a stereo-based end-to-end 3D object detection pipeline (Figure \ref{fig:small_pipeline}) -- Deep Stereo Geometry Network (DSGN), which relies on space transformation from 2D features to an effective 3D structure, called 3D geometric volume (3DGV). 

The insight behind 3DGV lies in the approach to construct the 3D volume that encodes 3D geometry. 3D geometric volume is defined in 3D world space, transformed from a plane-sweep volume (PSV) \cite{planesweep,deepstereo} constructed in the camera frustum. The pixel-correspondence constraint can be well learned in PSV, while 3D features for real-world objects can be learned in 3DGV. The volume construction is fully differentiable and thus can be jointly optimized for learning of both stereo matching and object detection.

This volumetric representation has two key advantages. First, it is easy to impose the pixel-correspondence constraint and encode full depth information into 3D real-world volume. Second, it provides 3D representation with geometry information that makes it possible to learn 3D geometric features for real-world objects. As far as we know, there was no study yet to explicitly investigate the way of encoding 3D geometry into an image-based detection network. Our contribution is summarized as follows.\vspace{-0.05in}

\begin{itemize}
   \item To bridge the gap between 2D image and 3D space, we establish stereo correspondence in a plane-sweep volume and then transform it to 3D geometric volume for capability to encode both 3D geometry and semantic cues for prediction in 3D regular space.\vspace{-0.05in}
   \item We design an end-to-end pipeline for extracting \textit{pixel-level} features for stereo matching and \textit{high-level} features for object recognition. The proposed network jointly estimates scene depth and detects 3D objects in 3D world, enabling many practical applications.\vspace{-0.05in}
   \item Without bells and whistles, our simple and fully-differentiable network outperforms all other stereo-based 3D object detectors (10 points higher in terms of AP) on the official KITTI leaderboard \cite{kitti}.
\end{itemize} 

\section{Related Work}

We briefly review recent work on stereo matching and multi-view stereo. Then we survey 3D object detection based on LiDAR, monocular images, and stereo images.

\vspace{0.1in}
\noindent{\bf Stereo Matching~~}  In the field of stereo matching on binocular images, methods of \cite{gcnet, psmnet, ganet, groupwisestereo, pwcnet, anytimestereo} process the left and right images by a Siamese network and construct a 3D cost volume to compute the matching cost. Correlation-based cost volume is applied in recent work \cite{dispnet, hierarchical, segstereo, groupwisestereo,learndispconsistency,edgestereo}. GC-Net \cite{gcnet} forms a concatenation-based cost volume and applies 3D convolution to regress disparity estimates. Recent PSMNet \cite{psmnet} further improves the accuracy by introducing pyramid pooling module and stacks hourglass modules \cite{hourglass}. %Learning-based approaches enhance the matching performance, which can introduce context cue for textureless regions and semantic cues like reflective and specular prior knowledge. 
State-of-the-art methods already achieved less than 2\% 3-pixel error on KITTI 2015 stereo benchmark. 

\vspace{0.1in}
\noindent{\bf Multi-View Stereo~~} Methods of \cite{point-mvsnet,mvsnet,surfacenet,MVSMachine,dpsnet,deepmvs} reconstruct 3D objects in a multi-view stereo setting \cite{MVSdata-dtu,shapenet2015}. 

MVSNet \cite{mvsnet} constructs plane-sweep volumes upon a \textit{camera frustum} to generate the depth map for each view. Point-MVSNet \cite{point-mvsnet} instead intermediately transforms the plane-sweep volume to point cloud representation to save computation. 
%However, a point cloud is much harder to be a differentiable operation and to normalize the coordinate for multiple objects, especially for a large scenario. 
Kar \etal \cite{MVSMachine} proposed the differentiable projection and unprojection operation on multi-view images. % The difference with our approach is that they learn a geometric constraint on 3D voxel occupancy grid (unprojected from 2D image) but our voxel representation is transformed from 3D cost volume that encodes the geometric constraint. 
%We do comparison experiment in ablation \ref{section:ablation}. 
% DeepMVS : It is unclear how the volumetric algorithms can be generalized and applied to high-resolution stereo reconstruction in the real-world.
%MVSNet: MVS Reconstruction. According to output representations, MVS methods can be categorized into 1) direct point cloud reconstructions [22,7], 2) volumetric reconstructions [20,33,14,15] and 3) depth map reconstructions [35,3,8,32,38].

\vspace{0.1in}
\noindent{\bf LiDAR-based 3D Detection~~} LiDAR sensors are very powerful, proven by several leading 3D detectors. Generally two types of architectures, i.e., voxel-based approaches \cite{VoxelNet,MV3D,pointpillar,fastpointrcnn} and point-based approaches \cite{pointnet,pointnet++,pointrcnn,std,pointfusion}, were proposed to process point cloud. %Due to the high accuracy of depth captured by LiDAR, they only have to deal with the representation of point cloud and extract semantic information properly.

\vspace{0.1in}
\noindent{\bf Image-based 3D Detection~~} Another line of detection is based on images. Regardless of monocular- or stereo-based setting, methods can be classified into two types according to intermediate representation existence.

\vspace{0.05in}
\noindent \textit{3D detector with depth predictor}: the solution relies on 2D image detectors and depth information extraction from monocular or stereo images.  Stereo R-CNN$_{\rm Stereo}$ \cite{stereorcnn} formulates 3D detection into multiple branches/stages to explicitly resolve several constraints. We note that the keypoint constraint may be hard to generalize to other categories like \textsl{Pedestrian}, and the dense alignment for stereo matching directly operating raw RGB images may be vulnerable to occlusion.
 
MonoGRNet$_{Mono}$ \cite{qin2019monogrnet} consists of four subnetworks for progressive 3D localization and directly learning 3D information based solely on semantic cues. MonoDIS$_{\rm Mono}$ \cite{disentangling} disentangles the loss for 2D and 3D detection. It achieves both tasks in an end-to-end manner. M3D-RPN$_{\rm Mono}$ \cite{brazil2019m3d} applies multiple 2D convolutions of non-shared weights to learn location-specific features for joint prediction of 2D and 3D boxes.
%MLF$_{Mono}$ \cite{MLF} fuses a coordinate map generated by another network extracting the RoI feature to regress the final 3D box. 
Triangulation$_{\rm Stereo}$ \cite{triangulation} directly learns offset from predefined 3D anchors on bird's eye view and establishes object correspondence on RoI-level features. Due to low resolutions, pixel correspondence is not fully exploited.
%Besides, all of the 2D-based 3D detectors have to face the problem of object occlusion and truncation in the image. 

\vspace{0.05in}
\noindent 
\textit{3D representation based 3D Detector}: 3DOP$_{\rm Stereo}$ \cite{3dop,3dop-pami} generates point cloud by stereo and encodes the prior knowledge and depth in an energy function. Several methods \cite{pseudolidar, pseudo++, monopseudoliar,accuratemono3d} transform the depth map to Pseudo-LiDAR (point cloud) intermediately followed by another independent network. This pipeline yields large improvement over previous methods. % Color-Embeding$_{Mono}$ \cite{accuratemono3d} extracts point cloud for each candidate boxes and PointNet \cite{pointnet} is used for predicting 3D boxes. %Pseudo-LiDAR$_{Stereo, Mono}$ \cite{pseudolidar, pseudo++} and AM3D$_{Mono}$ \cite{accuratemono3d} states that Pseudo-LiDAR as the middle representation is the key to the accuracy improvement. 
%However, this method is too cumbersome with several networks and the connection between stages is non-differentiable. Besides, 
%This representation encounters a streaking artifact problem that has been investigated by \cite{depthcoefficients, monopseudoliar, pseudo++}. 
OFT-Net$_{\rm Mono}$ \cite{oftnet} maps image feature into an orthographic bird's eye view representation and detects 3D objects on bird's eye view. 

% OC-stereo \cite{ocstereo} follows Pseudo-LiDAR and tries to focus on the foreground point cloud extracted by a 2D detector at first. The overall framework has several stages including detection, segmentation and then extracts the point cloud for 3D object detection. ROI-based stereo ???? can have lower resolution while indeed stereo matching requires finer resolution !!!, 

% ROI-based method requires accurate location from the left ROI and right ROI for stereo matching. OC-stereo matches the ROI-pair combinations by computing their Structural SIMilarity index (SSIM). Note that instance segmentation mask has to be used to remove the background disparity, which causes discontinuity of disparity.

\section{Our Approach}
In this section, we first explore the proper representation for 3D space and motivate our network design. Based on the discussion, we present our complete 3D detection pipeline under a binocular image pair setting.

\begin{figure*}
    \begin{center}
        \includegraphics[width=175mm]{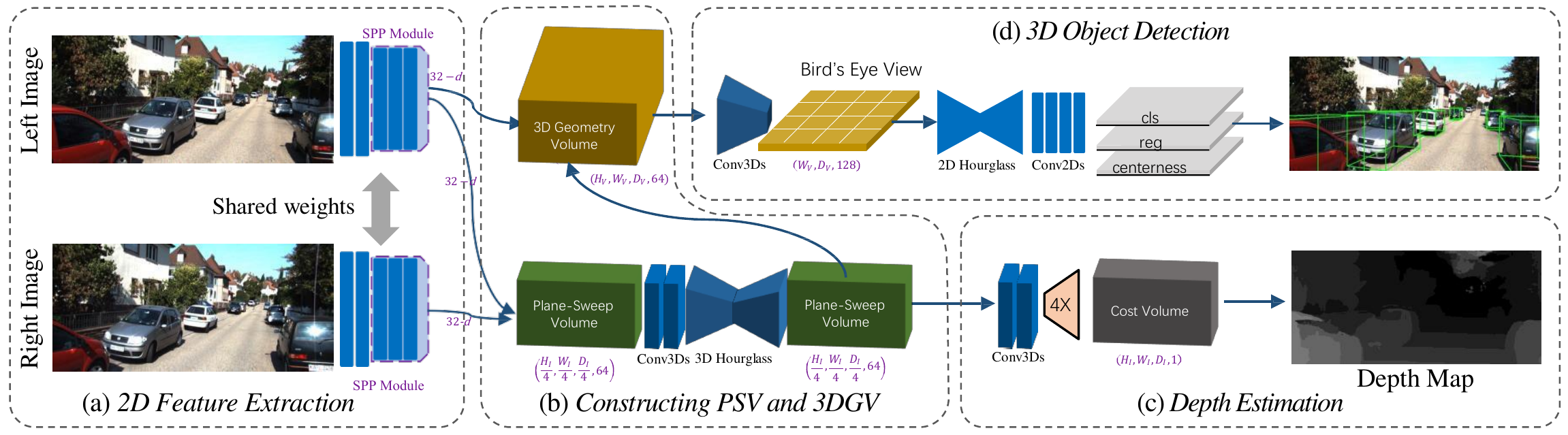}
    \end{center}\vspace{-0.1in}
    \caption{Overview of Deep Stereo Geometry Network (DSGN). The whole neural network consists of four components. (a) A 2D image feature extractor for capture of both \textit{pixel-} and \textit{high-level} feature. (b) Constructing the plane-sweep volume and 3D geometric volume. (c) Depth Estimation on the plane-sweep volume. (d) 3D object detection on 3D geometric volume. }
    \label{fig:pipeline}
\end{figure*}

\subsection{Motivation}\label{motivation}

Due to perspective, objects appear smaller with the increase of distance, which makes it possible to roughly estimate the depth according to the relative scale of objects sizes and the context. However, 3D objects of the same category may still have various sizes and orientations. It greatly increases the difficulty to make accurate prediction. 

Besides, the visual effect of \textit{foreshortening} causes that nearby 3D objects are not scaled evenly in images. A regular cuboid car appears like an irregular frustum. These two problems impose major challenges for 2D neural networks to model the relationship between 2D imaging and real 3D objects \cite{stereorcnn}. Thus, instead of relying on 2D representation, by reversing the process of projection, an intermediate 3D representation provides a more promising way for 3D object understanding. The following two representations can be typically used in 3D world.

\vspace{0.05in}
\noindent \textbf{Point-based Representation~~} Current state-of-the-art pipelines \cite{pseudolidar,pseudo++,accuratemono3d} generate intermediate 3D structure of point cloud by depth prediction approaches \cite{dorn,psmnet,gcnet} and apply LiDAR-based 3D object detectors. The main possible weakness is that it involves several independent networks and potentially loses information during intermediate transformation, making the 3D structure (such as cost volume) boiled down to point cloud. 

This representation often encounters streaking artifacts near object edges \cite{depthcoefficients, monopseudoliar, pseudo++}. Besides, the network is hard to be differentiated for multi-object scenes \cite{point-mvsnet,fastpointrcnn}.

\vspace{0.05in}
\noindent \textbf{Voxel-based Representation~~} Volumetric representation, as another way of 3D representation, is investigated less intensively. OFT-Net$_{\rm mono}$ \cite{oftnet} directly maps the image feature to the 3D voxel grid and then collapses it to the feature on bird's eye view. However, this transformation keeps the 2D representation for this view and does not explicitly encode the 3D geometry of data. 

% For image-based approach, we defines two geometric constraints: projection constraint and pixel-correspondence constraint. The former one is the correspondence of 3D and 2D features (many-to-one). The latter one is the pixel-correspondence of stereo image pair. The learning of depth enables the network to learn these two constraints from images. However, accurate depth information is important for 3D object detection in a larger scene.

\vspace{0.05in}
\noindent \textbf{Our Advantage~~} The key to establishment of an effective 3D representation relies on the ability to encode accurate 3D geometric information of the 3D space. A stereo camera provides an explicit pixel-correspondence constraint for computing depth. Aiming to design a unified network to exploit this constraint, we explore deep architectures capable of extracting both \textit{pixel-level} features for stereo correspondence and \textit{high-level} features for semantic cues. 

On the other hand, the pixel-correspondence constraint is supposedly imposed along the projection ray through each pixel where the depth is considered to be \textit{definite}. To this end, we create an intermediate plane-sweep volume from a binocular image pair to learn stereo correspondence constraint in camera frustum and then transform it to a 3D volume in 3D space. In this 3D volume with 3D geometric information lifted from the plane-sweep volume, we are able to well learn 3D features for real-world objects.

% \noindent \textbf{Feature Reuse} Instead of applying a new network \cite{pseudolidar,pseudo++,accuratemono3d}, network should be exploited to ease the computation through feature reuse. Only features of 32 dims are fed to cost volume in recent stereo network \cite{psmnet} so that it can limit the network capability in the following layers.

% \noindent \textbf{Discriminative Object Detector} Since the voxel feature is transformed from the subsampled feature map, some features might be over-sampled, which gives the challenge to learn a discriminative 3D object detector. \ylchen{to be proved in supplementary file}
% of the imbalanced positive and negative examples, the target assignment strategies are designed to increase the discriminative ability of the network. 

\subsection{Deep Stereo Geometry Network}
In this subsection, we describe our overall pipeline -- Deep Stereo Geometry Network (DSGN) as shown in Figure \ref{fig:pipeline}. Taking the input of a binocular image pair ($I_L, I_R$), we extract features by a Siamese network and construct a plane-sweep volume (PSV). The pixel-correspondence is learned on this volume. By differentiable warping, we transform PSV to a 3D geometric volume (3DGV) to establish 3D geometry in \textit{3D world space}. Then the following 3D neural network on the 3D volume learns necessary structure for 3D object detection.

\subsubsection{Image Feature Extraction}

Networks for stereo matching \cite{gcnet,psmnet,groupwisestereo} and object recognition \cite{resnet,vgg} have different architecture designs for their respective tasks. To ensure reasonable accuracy of stereo matching, we adopt the main design of PSMNet \cite{psmnet}. 
%For example, recent stereo matching networks \cite{psmnet,groupwisestereo} downsample the feature map by two or four but typical detection networks \cite{fasterrcnn, resnet} downsample the feature map by 16 or 32. Also, they apply average pooling instead of max-pooling and use basic residual block instead of Bottleneck module \cite{resnet}. 

Because the detection network requires a discriminative feature based on high-level semantic features and large context information, we modify the network for grasping more high-level information. Besides, the following 3D CNN for cost volume aggregation takes much more computation, which gives us room to modify the 2D feature extractor without introducing extra heavy computation overhead in the overall network. 

\paragraph{Network Architecture Details~~} Here we use the notations \texttt{conv\_1}, \texttt{conv\_2}, ..., \texttt{conv\_5} following \cite{resnet}. The key modification for 2D feature extractor is as follows. 

\begin{itemize}
	\item Shift more computation from \texttt{conv\_3} to \texttt{conv\_4} and \texttt{conv\_5}, \ie, changing the numbers of basic blocks of \texttt{conv\_2} to \texttt{conv\_5} from $\{3, 16, 3, 3\}$ to $\{3, 6, 12, 4\}$.\vspace{-0.05in}
	\item The SPP module used in PSMNet concatenates the output layers of \texttt{conv\_4} and \texttt{conv\_5}.\vspace{-0.05in}
	\item The output channel number of convolutions in \texttt{conv\_1} is $64$ instead of $32$ and the output channel number of a basic residual block is $192$ instead of $128$.
\end{itemize} 

Full details of our 2D feature extraction network are included in the supplementary material.

\subsubsection{Constructing 3D Geometric Volume}

To learn 3D convolutional features in 3D regular space, we first create a 3D geometric volume (3DGV) by warping a plane-sweep volume to 3D regular space. Without loss of generality, we discretize the region of interest in \textit{3D world space} to a 3D voxel occupancy grid of size $(W_V, H_V, D_V)$ along the right, down and front directions in camera view. $W_V, H_V, D_V$ denote the width, height and length of the grid, respectively. Each voxel is of size $(v_w, v_h, v_d)$.  

\vspace{0.05in}
\noindent \textbf{Plane-Sweep Volume~~} In binocular vision, an image pair $(I_L, I_R)$ is used to construct a disparity-based cost volume for computing matching cost, which matches a pixel $i$ in the left image $I_L$ to the correspondence in the right image $I_R$ horizontally shifted by an integral disparity value $d$. The depth is {inversely} proportional to disparity. %:
% \begin{equation}
%    \mathit{depth} = \frac{f_x \times b}{\mathit{disparity}},
% \end{equation}
% where $f_x$ and $b$ are the known horizontal focal length and baseline. 

It is thus hard to distinguish among distant objects due to the similar disparity values \cite{stereorcnn,pseudolidar,pseudo++}. For example, objects $40$-meter and $39$-meter away have almost no difference ($<0.25$pix) on disparity on KITTI benchmark \cite{kitti}. %Cost volume in this paper is referred to the volume formed to compute matching cost but its implementation in stereo matching \cite{dispnet,gcnet,psmnet}. 

In a different way to construct the cost volume, we follow the classic plane sweeping approach \cite{planesweep,deepstereo,mvsnet} to construct a plane-sweep volume by concatenating the left image feature $F_L$ and the reprojected right image feature $F_{R->L}$ at equally spaced depth interval, which avoids imbalanced mapping of features to 3D space. 

The coordinate of PSV is represented by $(u, v, d)$, where $(u, v)$ represents $(u,v)$-pixel in the image and it adds another axis orthogonal to the image plane for depth. We call the space of $(u, v, d)$ grid \textit{camera frustum space}. The depth candidates $d_i$ are uniformly sampled along the depth dimension with interval $v_d$ following the pre-defined 3D grid. Concatenation-based volume enables the network to learn semantic features for object recognition. 

We apply 3D convolution to this volume and finally get a matching cost volume for all depth. To ease computation, we apply only one 3D hourglass module, contrary to the three used in PSMNet \cite{psmnet}. We note that the resulting performance degradation can be compensated in the following detection network since the overall network is differentiable.

\begin{figure}
    \includegraphics[width=86mm]{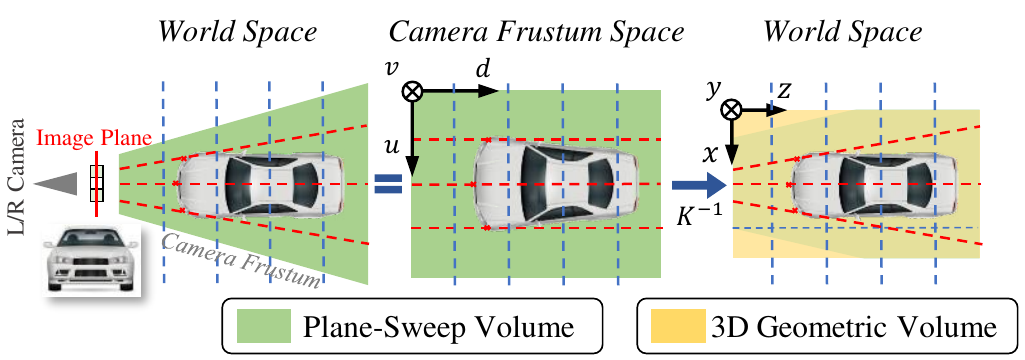}
    \caption{Illustration of volume transformation. The image is captured at the image plane (red solid line). PSV is constructed by projecting images at equally spaced depth (blue dotted lines) in left camera frustum, which is shown in the \textit{3D world space} (left) and \textit{camera frustum space} (middle). Car is shown to be distorted in the middle. Mapping by the camera intrinsic matrix $K$, PSV is warped to 3DGV, which restores the car.}
    \label{fig:transformation}\vspace{-0.1in}
\end{figure}

\vspace{0.05in}
\noindent \textbf{3D Geometric Volume~~} 
With known camera internal parameters, we transform the last feature map of PSV before computing matching cost from \textit{camera frustum space} $(u,v,d)$ to \textit{3D world space} $(x,y,z)$ by reversing 3D projection with
\begin{equation}\label{project}
\begin{split}
\begin{pmatrix}
x\\y\\z
\end{pmatrix}
=
\begin{pmatrix}
1/f_x & 0 & -c_u/f_x \\
0 & 1/f_y & -c_v/f_y \\
0 & 0 & 1
\end{pmatrix}
\begin{pmatrix}
ud\\vd\\d
\end{pmatrix}
\end{split}
\end{equation}
where $f_x, f_y$ are the 
horizontal and vertical focal lengths. This transformation is fully-differentiable and saves computation by eliminating background outside the pre-defined grid, such as the sky. It can be implemented by warp operation with \textit{trilinear} interpolation.

Figure \ref{fig:transformation} illustrates the transformation process. %, it is hard to impose constraint along the slanting ray on 3D volume in 3D world (red lines in rightmost subfigure) but the ray is parallel to the voxel in plane-sweep volume (red lines in middle subfigure). The insight here is 
The common pixel-correspondence constraint (red dotted lines) is imposed in \textit{camera frustum} while object recognition is learned in regular \textit{3D world space} (\textit{Euclidean space}). There obviously is difference in these two representations.

In the last feature map of plane-sweep volume, a low-cost voxel $(u, v, d)$ means the high probability of object existing at depth $d$ along the ray through the focal point and image point $(u,v)$. With the transformation to regular \textit{3D world space}, the feature of low cost suggests that this voxel is occupied in the front surface of the scene, which can serve as a feature for 3D geometric structure. Thus it is possible for the following 3D network to learn 3D object features on this volume.

This operation is fundamentally different from differentiable unprojection \cite{MVSMachine}, which directly lifts the image feature from 2D image frame to 3D world by \textit{bilinear} interpolation. Our goal is to lift geometric information from cost volume to 3D world grid. We make pixel-correspondence constraint easy to be imposed along the projection ray. 

The contemporary work \cite{pseudo++} applies a similar idea to construct depth-cost-volume like plane-sweep volume. Differently, we aim to avoid imbalanced warping from plane-sweep volume to 3D geometric volume, and deal with the streaking artifact problem. Besides, our transformation keeps the distribution of depth instead of deducting it to a depth map. Our strategy intriguingly avoids object artifacts.

\subsubsection{Depth Regression on Plane-Sweep Cost Volume}\label{stereo matching on PSV}
To compute the matching cost on the plane-sweep volume, we reduce the final feature map of plane-sweep volume by two 3D convolutions to get 1D cost volume (called plane-sweep cost volume). {Soft arg-min} operation \cite{gcnet, psmnet, ganet} is applied to compute the expectation for all depth candidates with probability $\sigma(-c_d)$ as
\begin{equation}
\hat{d} = \sum_{d\in \{z_\text{min}, z_\text{min} + v_d, ..., z_\text{max}\}} d \times \sigma(-c_d)
\end{equation}
where the depth candidates are uniformly sampled within pre-defined grid $[z_\text{min}, z_\text{max}]$ with interval $v_d$. The {softmax} function encourages the model to pick a single depth plane per pixel.

\subsubsection{3D Object Detector on 3D Geometric Volume}

Motivated by recent one-stage 2D detector FCOS \cite{fcos}, we extend the idea of \textit{centerness} branch in our pipeline and design a distance-based strategy to assign targets for the real world. Because objects of the same category are of similar size in 3D scene, we still keep the design of anchors. 

Let $\mathcal{V} \in \mathbb{R}^{W\times H\times D\times C}$ be the feature map for 3DGV of size $(W, H, D)$ and denote the channels as $C$. Considering the scenario of autonomous driving, we gradually downsample along the height dimension and finally get the feature map $\mathcal{F}$ of size $(W, H)$ for bird's eye view. The network architecture is included in the supplementary material.

For each location $(x, z)$ in $\mathcal{F}$, several anchors of different orientations and sizes are placed. Anchors $\mathbf{A}$ and ground-truth boxes $\mathbf{G}$ are represented by the location, prior size and orientation, \ie, ($x_\mathbf{A}, y_\mathbf{A}, z_\mathbf{A}, h_\mathbf{A}, w_\mathbf{A}, l_\mathbf{A}, \theta_A$) and ($x_\mathbf{G}, y_\mathbf{G}, z_\mathbf{G}, h_\mathbf{G}, w_\mathbf{G}, l_\mathbf{G}, \theta_\mathbf{G}$). Our network regresses from anchor and gets the final prediction 
$(h_\mathbf{A} e^{\delta h}, w_\mathbf{A} e^{\delta w}, l_\mathbf{A} e^{\delta l}, x_\mathbf{A} + \delta x, y_\mathbf{A} + \delta y, z_\mathbf{A} + \delta z, \theta_\mathbf{A} + \pi / N_\theta \tanh(\delta \theta))$,
where $N_\theta$ denotes the number of anchor orientations and $\delta \cdot$ is the learned offset for each parameter.

\vspace{0.05in}
\paragraph{Distance-based Target Assignment}
Taking object orientation into consideration, we propose distance-based target assignment. The distance is defined as the distance of 8 corners between anchor and ground-truth boxes as
\begin{equation}
distance(\mathbf{A}, \mathbf{G}) = \frac{1}{8}\sum_{i=1}^8{\sqrt{(x_{\mathbf{A}_i} - x_{\mathbf{G}_i})^2 + (z_{\mathbf{A}_i} - z_{\mathbf{G}_i})^2)}}   \nonumber
\end{equation}

In order to balance the ratio of positive and negative samples, we let the anchors with top $N$ nearest distance to ground-truth as positive samples, where $N = \gamma \times k$ and $k$ is the number of voxels inside ground-truth box on bird's eye view. $\gamma$ adjusts the number of positive samples. Our centerness is defined as the exponent of the negative normalized distance of eight corners as
\begin{equation}
\begin{split}
centerness(\mathbf{A},\mathbf{G}) = e^{-\text{norm}(distance(\mathbf{A},\mathbf{G}))}, 
\end{split}  
\end{equation}
where $\text{norm}$ denotes min-max normalization.
% \begin{equation}
% \text{norm}(X) = \frac{X - \min(X)}{\max(X) - \min(X)}
% \end{equation}

% We note that \cite{disentangling} proposes a similar definition of 3D bounding box confidence. But they only use it as the posterior confidence of 3D bounding box given 2D proposal but we use it also in the training \cite{fcos}. Besides, two other target assignment strategies are discussed in supplementary material.

\subsection{Multi-task Training}\label{multi-task training}
Our network with stereo matching network and 3D object detector is trained in an end-to-end fashion. We train the overall 3D object detector with a multi-task loss as 
\begin{equation}
Loss = \mathcal{L}_{depth} + \mathcal{L}_{cls} + \mathcal{L}_{reg} + \mathcal{L}_{centerness}.
\end{equation}
For the loss of depth regression, we adopt {smooth $L_1$ loss} \cite{gcnet} in this branch as
\begin{equation}
\mathcal{L}_{depth} = \frac{1}{N_D}\sum_{i=1}^{N_D} smooth_{L_1}\left(d_i - \hat{d_i}\right),
\end{equation}
where $N_D$ is the number of pixels with ground-truth depth (obtained from the sparse LiDAR sensor).

For the loss of classification, {focal loss} \cite{focalloss} is adopted in our network to deal with the class imbalance problem in 3D world as
\begin{equation}
\mathcal{L}_{cls} = \frac{1}{N_{pos}} \sum_{(x,z)\in \mathcal{F}} \textit{Focal\_Loss}(p_{\mathbf{A}_{(x,z)}}, p_{\mathbf{G}_{(x,z)}}),
\end{equation}
where $N_{pos}$ denotes the number of positive samples. {Binary cross-entropy (BCE) loss} is used for centerness.

For the loss of 3D bounding box regression, {smooth $L_1$ Loss} is used for the regression of bounding boxes as
\begin{equation}
\begin{split}
\mathcal{L}_{reg} = \frac{1}{N_{pos}}\sum_{(x,z)\in F_{pos}} {\rm centerness}(\mathbf{A},\mathbf{G}) \times\\
{\rm smooth}_{L_1}( l1\_{\rm distance}(\mathbf{A}, \mathbf{G}) )
\end{split}
\end{equation}
where $F_{pos}$ denotes all positive samples on bird's eye view.

We try two different regression targets with and without jointly learning all parameters. 

\begin{itemize}
	\item {\textit{Separably optimizing box parameters}}. The regression loss is directly applied to the offset of ($x, y, z, h, w, l, \theta$).\vspace{-0.05in}
	\item {\textit{Jointly optimizing box corners}}. For jointly optimizing box parameters, the loss is made on the average \textit{L1} distance of eight box corners between predicted boxes from 3D anchors and ground-truth boxes following that of \cite{fpointnet}.
\end{itemize}

% The \textit{Sine-Error Loss} proposed by SECOND \cite{second} is used for angle regression, but we make loss directly on orientation by multiplying an extra coefficient $2$ instead of an extra direction classifier for orientation. \ie.,
% % \begin{equation}
% % Sin\_distance_\theta(\mathbf{A}, \mathbf{G}) = sin(\theta_\mathbf{A} - \theta_\mathbf{G})
% % \end{equation}
% \begin{equation}
% Sin\_distance_\theta(\mathbf{A}, \mathbf{G}) = sin(2\theta_\mathbf{A} - 2\theta_\mathbf{G})
% \end{equation}
% Note that other parameters except $\theta$ directly apply \textit{Smooth L1 Loss}.

% \ie.
% \begin{equation}
% \begin{split}
% &l1\_distance(\mathbf{A}, \mathbf{G}) = \\
% &\frac{1}{8}\sum_{i=1}^8\left({|x_{\mathbf{A}_i} - x_{\mathbf{G}_i}| + |y_{\mathbf{A}_i} - y_{\mathbf{G}_i}| + |z_{\mathbf{A}_i} - z_{\mathbf{G}_i}|}\right)
% \end{split}
% \end{equation}

In our experiments, we use the second regression target for \textsl{Car} and the first regression target for \textsl{Pedestrian} and \textsl{Cyclist}. Because it is hard for even human to accurately predict or annotate the orientation of objects like \textsl{Pedestrian} from an image, other parameter estimation under joint optimization can be affected.

\section{Experiments}

\noindent \textbf{Datasets~~}
Our approach is evaluated on the popular KITTI 3D object detection dataset \cite{kitti}, which is union of $7,481$ stereo image-pairs and point clouds for training and $7,518$ for testing. The ground-truth depth maps are generated from point clouds following \cite{pseudolidar,pseudo++}. The training data has annotation for \textsl{Car}, \textsl{Pedestrian} and \textsl{Cyclist}. The KITTI leaderboard limits the access to submission to the server for evaluating test set. Thus, following the protocol in \cite{MV3D, stereorcnn, pseudolidar}, the training data is divided into a training set (3,712 images) and a validation set (3,769 images). All ablation studies are conducted on the split. For the submission of our approach, our model is trained from scratch on the 7K training data only. 

\vspace{0.05in}
\noindent \textbf{Evaluation Metric~~}\label{evaluation metric}
KITTI has three levels of difficulty setting of easy, moderate (main index) and hard, according to the occlusion/truncation and the size of an object in the 2D image. All methods are evaluated for three levels of difficulty under different IoU criteria per class , \ie, IoU $\geq$ 0.7 for \textsl{Car} and IoU $\geq$ 0.5 for \textsl{Pedestrian} and \textsl{Cyclist} for 2D, bird's eye view and 3D detection. 

Following most image-based 3D object detection setting \cite{pseudolidar, stereorcnn, triangulation,qin2019monogrnet,brazil2019m3d,disentangling}, the ablation experiments are conducted on \textsl{Car}. We also report the results of \textsl{Pedestrian} and \textsl{Cyclist} for reference in the supplementary file. KITTI benchmark recently changes evaluation where AP calculation uses 40 recall positions instead of the 11 recall positions proposed in the original Pascal VOC benchmark. Thus, we show the main test results following the official KITTI leaderboard. We generate the validation results using the original evaluation code for fair comparison with other approaches in ablation studies.

\subsection{Implementation}\label{section:implementation}

\noindent \textbf{Training Details~~}
By default, models are trained on 4 \textit{NVIDIA Tesla V100} (32G) GPUs with batch-size 4 -- that is, each GPU holds one pair of stereo images of size $384\times 1248$. 
%For reference, PSMNet \cite{psmnet} trains the network with $256\times 512$ patch with 3 pairs each GPU. 
We apply \textit{ADAM} \cite{adam} optimizer with initial learning rate 0.001. We train our network for 50 epochs and the learning rate is decreased by 10 at 50-th epoch. The overall training time is about 17 hours. The data augmentation used is horizontal flipping only. 

Following other approaches \cite{pseudo++,pseudolidar,VoxelNet,second,pointrcnn,fastpointrcnn}, another network is trained for \textsl{Pedestrian} and \textsl{Cyclist}, we first pre-train the network with all training images for the stereo network and then apply fine-tune with 3D box annotation for both branches because only about $1/3$ images have annotations of these two objects. 

\noindent \textbf{Implementation Details~~}
For constructing plane-sweep volume, the image feature map is shrunk to 32D and downsampled by 4 for both left and right images. Then by reprojection and concatenation, we construct the volume of shape $(W_{I}/4, H_{I}/4, D_{I}/4, 64)$, where the image size is $(W_{I}=1248, H_{I}=384)$ and the number of depth is $D_{I}=192$. It is followed by one 3D hourglass module \cite{psmnet, hourglass} and extra 3D convolutions to get the matching cost volume of shape $(W_{I}/4, H_{I}/4, D_{I}/4, 1)$. Then interpolation is used to upsample this volume to fit the image size. 

To construct 3D geometric volume, We discretize the region in range $[-30.4, 30.4]\times [-1, 3]\times [2, 40.4]$ (meters) to a 3D voxel occupancy grid of size $(W_{V}=300, H_{V}=20, D_{V}=192)$ along the right ($X$), down ($Y$) and front ($Z$) directions in camera's view. 3D geometric volume is formed by warping the last feature map of PSV. Each voxel is a cube of size $(0.2, 0.2, 0.2)$ (meter). 

Other implementation details and the network architecture are included in the supplementary file.

\subsection{Main Results}

\begin{table*}[bpt]\footnotesize
\begin{center}
\begin{tabular}{c|c|c|c|c|c|c|c|c|c|c}
\specialrule{1pt}{1pt}{1pt}
\multirow{2}{*}{Modality} & \multirow{2}{*}{Method} & \multicolumn{3}{c|}{ \textbf{3D Detection AP} (\textbf{\%})} & \multicolumn{3}{c|}{BEV Detection AP (\%)} & \multicolumn{3}{c}{2D Detection AP (\%)} \\ \cline{3-11}
& & Easy & \textbf{Moderate} & Hard & Easy & \textbf{Moderate} & Hard & Easy & \textbf{Moderate} & Hard \\ \hline \hline
\multirow{1}{*}{LiDAR}
 & MV3D (LiDAR) \cite{MV3D} & 68.35 & 54.54 & 49.16 & 86.49 & 78.98 & 72.23 & -- & -- & -- \\
% & PL++: P-RCNN +SL \cite{pseudo++} & 68.38 & 54.88 & 49.16 & 84.61 & 73.80 & 65.59 & 94.95 & 85.15 & 77.78 \\  
\hline \hline
\multirow{4}{*}{Mono} & OFT-Net \cite{oftnet} & 1.61 & 1.32 & 1.00 & 7.16 & 5.69 & 4.61 & -- & -- & --  \\ 
& MonoGRNet \cite{qin2019monogrnet} & 9.61 & 5.74 & 4.25 & 18.19 & 11.17 & 8.73 & 88.65 & 77.94 & 63.31 \\
& M3D-RPN \cite{brazil2019m3d} & 14.76 & 9.71 & 7.42 & 21.02 & 13.67 & 10.23 & 89.04 & 85.08 & 69.26 \\
& AM3D \cite{accuratemono3d} & 16.50 & 10.74 & 9.52 & 25.03 & 17.32 & 14.91 & 92.55 & {\bf 88.71} & 77.78 \\
\hline 
\multirow{4}{*}{Stereo} 
& 3DOP \cite{3dop} & -- & -- & -- & -- & -- & -- & 93.04 & 88.64 & {\bf 79.10} \\
& Stereo R-CNN* \cite{stereorcnn} & 47.58 & 30.23 & 23.72 & 61.92 & 41.31 & 33.42 & 93.98 & 85.98 & 71.25 \\ 
& PL: AVOD* \cite{pseudolidar} & 54.53 & 34.05 & 28.25 & 67.30 & 45.00 & 38.40 & 85.40 & 67.79 & 58.50 \\ 
& PL++: P-RCNN* \cite{pseudo++} & 61.11 & 42.43 & 36.99 & 78.31 & 58.01 & 51.25 & 94.46 & 82.90 & 75.45 \\  
& DSGN (Ours) & {\bf 73.50} & {\bf 52.18} & {\bf 45.14} & {\bf 82.90} & {\bf 65.05} & {\bf 56.60} & {\bf 95.53} & 86.43 & 78.75 \\
\specialrule{1pt}{1pt}{1pt}
\end{tabular}
\end{center}
\caption{Comparison of main results on KITTI test set (official KITTI leaderboard). The results are evaluated using new evaluation metric on the KITTI leaderboard. Several methods undergoing old evaluation are not available on the leaderboard.
PL/PL++* uses extra Scene Flow dataset to pre-train the stereo matching network and Stereo R-CNN* uses ImageNet pre-trained model.}
\label{table: Comparison of main results on KITTI test set.}
\end{table*}

\begin{table*}[bpt]\footnotesize
\begin{center}
\begin{tabular}{c|c|c|c|c|c|c|c|c|c|c}
\specialrule{1pt}{1pt}{1pt}
\multirow{2}{*}{Modality} & \multirow{2}{*}{Method} & \multicolumn{3}{c|}{ \textbf{3D Detection AP} (\textbf{\%})} & \multicolumn{3}{c|}{BEV Detection AP (\%)} & \multicolumn{3}{c}{2D Detection AP (\%)}\\ \cline{3-11}
& & Easy & \textbf{Moderate} & Hard & Easy & \textbf{Moderate} & Hard & Easy & \textbf{Moderate} & Hard \\ \hline \hline
\multirow{1}{*}{LiDAR}
 & MV3D (LiDAR) \cite{MV3D} & 71.29 & 56.60 & 55.30 & 86.18 & 77.32 & 76.33 & 88.41 & 87.76 & 79.90 \\ 
% & PL++: P-RCNN +SL \cite{pseudo++} & 75.1 & 63.8 & 57.4 & 88.2 & 76.9 & 73.4 & -- & -- & -- \\ 
\hline \hline
\multirow{4}{*}{Mono} & OFT-Net \cite{oftnet} & 4.07 & 3.27 & 3.29 & 11.06 & 8.79 & 8.91 & -- & -- & -- \\
& MonoGRNet \cite{qin2019monogrnet} & 13.88 & 10.19 & 7.62 & 43.75 & 28.39 & 23.87 & -- & 78.14 & -- \\ 
& M3D-RPN \cite{brazil2019m3d} & 20.27 & 17.06 & 15.21 & 25.94 & 21.18 & 17.90 & 90.24 & 83.67 & 67.69 \\
& AM3D \cite{accuratemono3d} & 32.23 & 21.09 & 17.26 & 43.75 & 28.39 & 23.87 & -- & -- & -- \\ \hline 
\multirow{9}{*}{Stereo} & MLF \cite{MLF} & -- & 9.80 & -- & -- & 19.54 & -- & -- & -- & -- \\ 
& 3DOP \cite{3dop} & 6.55 & 5.07 & 4.10 & 12.63 & 9.49 & 7.59 & -- & -- & -- \\ 
& Triangulation \cite{triangulation} & 18.15 & 14.26 & 13.72 & 29.22 & 21.88 & 18.83 & -- & -- & -- \\
& Stereo R-CNN* \cite{stereorcnn} & 54.1 & 36.7 & 31.1 & 68.5 & 48.3 & 41.5 & {\bf 98.73} & {\bf 88.48} & 71.26 \\ 
& PL: F-PointNet* \cite{pseudolidar} & 59.4 & 39.8 & 33.5 & 72.8 & 51.8 & 33.5 & -- & -- & --  \\ 
& PL: AVOD* \cite{pseudolidar} & 61.9 & 45.3 & 39.0 & 74.9 & 56.8 & 49.0 & -- & -- & -- \\ 
% As we can see, PL++ (AVOD) only improves little compared with PL which can also shows that the depth cost volume might improve little for voxel-based method but the depth distribution actually improves a lot.
& PL++: AVOD* \cite{pseudo++} & 63.2 & 46.8 & 39.8 & 77.0 & 63.7 & 56.0 & -- & -- & -- \\
& PL++: PIXOR* \cite{pseudo++} & -- & -- & -- & 79.7 & 61.1 & 54.5 & -- & -- & -- \\
& PL++: P-RCNN* \cite{pseudo++} & 67.9 & 50.1 & 45.3 & 82.0 & {\bf 64.0} & 57.3 & -- & -- & -- \\
& DSGN (Ours) & {\bf 72.31} & {\bf 54.27} & {\bf 47.71} & {\bf 83.24} & 63.91 & {\bf 57.83} & 89.25 & 83.59 & {\bf 78.45}  \\ 
\specialrule{1pt}{1pt}{1pt}
\end{tabular}
\end{center}
\caption{Comparison of main results on KITTI \textit{val} set. As described in Section \ref{evaluation metric}, we use original KITTI evaluation metric here. 
%PL++: P-RCNN + SL means that it inputs sparse 4-beam LiDAR. 
PL/PL++* uses extra Scene Flow dataset to pre-train the stereo matching network and Stereo R-CNN* uses ImageNet pre-trained model.}
\label{table: Comparison of main results on KITTI val set.}
\end{table*}

We give comparison with state-of-the-art 3D detectors in Tables \ref{table: Comparison of main results on KITTI test set.} and \ref{table: Comparison of main results on KITTI val set.}. Without bells and whistles, our approach outperforms all other image-based methods on 3D and BEV object detection. We note that Pseudo-LiDARs \cite{pseudolidar,pseudo++} is with pre-trained PSMNet \cite{psmnet} on a large-scale synthetic scene flow dataset \cite{dispnet} (with 30,000+ pairs of stereo images and dense disparity maps) for stereo matching. Stereo R-CNN \cite{stereorcnn} uses ImageNet pre-trained ResNet-101 as backbone and has input images of resolution $600\times 2000$. 

Differently, our model is trained from scratch only on these 7K training data with input of  resolution $384\times 1248$. Also, Pseudo-LiDARs \cite{pseudolidar,pseudo++} approaches apply two independent networks including several LiDAR-based detectors, while ours is just one unified network.

DSGN without explicitly learning 2D boxes surpasses those applying strong 2D detectors based on ResNet-101 \cite{stereorcnn} or DenseNet-121 \cite{brazil2019m3d}. It naturally achieves duplicate removal by non-maximum suppression (NMS) in 3D space, which coincides with the common belief that there is no collision between regular objects.
%We think it is due to the natural assumption that there is no collision between regular objects in 3D space. It provides a potential end-to-end approach for image-based object detection to overcome the problem of occlusion by transforming to 3D space.

More intriguingly, as shown in Table \ref{table: Comparison of main results on KITTI test set.}, DSGN even achieves comparable performance on BEV detection and better performance on 3D detection on KITTI easy regime with MV3D \cite{MV3D} (with LiDAR input only) -- a classic LiDAR-based 3D object detector, \textit{for the first time}. This result demonstrates a promising future application at least in the scenario of low-speed autonomous driving. 

The above comparison manifests the effectiveness of 3D geometric volume, which serves as a link between 2D images and 3D space by combining the depth information and semantic feature.

\vspace{0.05in}
\noindent \textbf{Inference Time~~}
On a {NVIDIA Tesla V100} GPU, the inference time of DSGN for one image pair is 0.682s on average, where 2D feature extraction for left and right images takes 0.113s, constructing the plane-sweep volume and 3D geometric volume takes 0.285s, and 3D object detection on 3D geometric volume takes 0.284s. The computation bottleneck of DSGN lies on 3D convolution layers. 

\subsection{Ablation Study}\label{section:ablation}

\subsubsection{Ablation study of 3D Volume Construction}

One of the main obstacles to construct an effective 3D geometric representation is the appropriate way of learning 3D geometry. We therefore investigate the effect of following three key components to construct a 3D volume.

\vspace{0.05in}
\noindent {\it Input Data~~}
Monocular-based 3D volume only has the potential to learn the correspondence between 2D and 3D feature, while stereo-based 3D volume can learn extra 2D feature correspondence for pixel-correspondence constraint. 

 \vspace{0.05in}
\noindent {\it Constructing 3D Volume~~}
One straightforward solution to construct 3D volume is by directly projecting the image feature to 3D voxel grid \cite{MVSMachine,oftnet} (denoted as IMG$\rightarrow$3DV). Another solution in Figure \ref{fig:transformation} transforms plane-sweep volume or disparity-based cost volume to 3D volume, which provides a natural way to impose pixel-correspondence constraint along the projection ray in camera frustum (denoted as IMG$\rightarrow$(PS)CV$\rightarrow$3DV). 

 \vspace{0.05in}
\noindent {\it Supervising Depth~~}
Supervised with or without the point cloud data, the network learns the depth explicitly or implicitly. One way is to supervise the voxel occupancy of 3D grid by ground-truth point cloud using \textit{binary cross-entropy loss}. The second is to supervise depth on the plane-sweep cost volume as explained in Section \ref{multi-task training}. 

For fair comparison, the models IMG$\rightarrow$3DV and IMG$\rightarrow$(PS)CV$\rightarrow$3DV have the same parameters by adding the same 3D hourglass module for the model IMG$\rightarrow$3DV. In addition, several important facts can be revealed from Table \ref{table: constraint} and are explained in the following. 

\begin{table}[t]\footnotesize
\begin{center}
\begin{tabular}{c|c|c|c}
\specialrule{1pt}{1pt}{1pt}
Input & Transformation & Supervision & AP$_{\text{3D}}$ / AP$_{\text{BEV}}$ / AP$_{\text{2D}}$ \\ \hline
\multirow{2}{*}{Mono} & \multirow{2}{*}{IMG$\rightarrow$3DV} & $\times$ & 6.22 / 11.98 / 58.23   \\ 
 & & 3DV & 13.66 / 19.92 / 65.89 \\ \hline
 % \cline{2-5} % & IMG$\rightarrow$PSCV$\rightarrow$3DV & \checkmark & $\times$ & 10.87 / 19.6  \\ \hline
\multirow{6}{*}{Stereo} 
 & \multirow{2}{*}{IMG$\rightarrow$3DV} & $\times$ & 11.03 / 15.17 / 57.30 \\ 
 & & 3DV & 42.57 / 54.55 / 81.86 \\ \cline{2-4}
 & IMG$\rightarrow$CV$\rightarrow$3DV & CV & 45.89 / 58.40 / 81.71 \\ \cline{2-4}
 & \multirow{2}{*}{IMG$\rightarrow$PSCV$\rightarrow$3DV} & $\times$ & 38.48 / 52.85 / 77.83 \\ 
 % & & 3DV & 56.62 / 67.29 \\ 
 & & PSCV & \textbf{54.27} / \textbf{63.91} / \textbf{83.59} \\ 
\specialrule{1pt}{1pt}{1pt}
\end{tabular}
\end{center}
\caption{Ablation study of depth encoded approaches. ``PSCV" and ``3DV" with ``Supervision'' header represent that the constraint is imposed in (plane-sweep) cost volume and 3D volume, respectively. The results are evaluated in moderate level.}
%\vspace{-0.1in}
\label{table: constraint}
\end{table}

\vspace{0.05in}
\noindent{\textit{Supervision of point cloud is important.~}}
The approaches under the supervision of the LiDAR point cloud consistently perform better than those without supervision, which demonstrates the importance of 3D geometry for image-based approaches. %The model IMG$_{stereo}$$\rightarrow$3DV with supervision on 3DV performs better than current state-of-the-art approaches PL:F-PointNet \cite{pseudolidar} and Stereo R-CNN \cite{stereorcnn}.

\vspace{0.05in}
\noindent{\textit{Stereo-based approaches work much better than monocular ones under supervision.~}}
The discrepancy between stereo and monocular approaches indicates that direct learning of 3D geometry from semantic cues is a quite difficult problem. In contrast, image-based approaches without supervision make these two lines yield similar performance, which indicates that supervision only by 3D bounding boxes is insufficient for learning of 3D geometry. 

\vspace{0.05in}
\noindent{\textit{Plane-sweep volume is a more suitable representation for 3D structure.~}}
Plane-sweep cost volume ($54.27$ AP) performs better than disparity-based cost volume ($45.89$ AP). It shows that balanced feature mapping is important during the transformation to 3D volume.

\vspace{0.05in}
\noindent{\textit{Plane-sweep volume, as an intermediate encoder, more effectively contains depth information.~}}
The inconsistency between IMG$\rightarrow$PSCV$\rightarrow$3DV and IMG$\rightarrow$3DV manifests that plane-sweep volume as the intermediate representation can effectively help learning of depth information. The observation explains that
the \textit{soft arg-min} operation encourages the model to pick a single depth plane per pixel along the projection ray, which shares the same spirit as the assumption that only one depth-value is true for each pixel. Another reason can be that PSCV and 3DV have different matching densities -- PSCV intermediately imposes the dense pixel correspondence over all image pixels. In contrast, only the left-right pixel pairs through the voxel centers are matched on 3DV.

From above comparison of volume construction, we observe that the three key facts affect the performance of computation pipelines. The understanding and recognition of how to construct a suitable 3D volume is still at the very early stage. More study is expected to reach comprehensive understanding of the volume construction from the multi-view images.
% The reason is that \text{softmax} function in cost volume encourages the model to pick a single depth plane per pixel, which shares the same spirit as the assumption that only one depth value (in the front surface) is true.

\subsubsection{Influence on Stereo Matching}

We conduct experiments for investigating the influence of depth estimation, which is evaluated on KITTI \textit{val} set following \cite{pseudolidar}. The average and median value of absolute depth estimation errors within the pre-defined range of $[z_\text{min}, z_\text{max}]$ is shown in Table \ref{table: stereo matching}. A natural baseline for our approach is PSMNet-PSV* modified from PSMNet \cite{psmnet} whose 2D feature extractor takes $0.041$s while ours takes $0.113$s.

Trained with depth estimation branch only, DSGN performs slightly better than PSMNet-PSV* with the same training pipeline in depth estimation.
For joint training of both tasks, both approaches suffer from larger and similar depth error ($0.5586$ meter for DSGN vs. $0.5606$ meter for PSMNet-PSV*). Differently, DSGN outperforms the alternatives by $7.86$ AP on 3D object detection and $6.34$ AP on BEV detection. The comparison indicates that our 2D network extracts better high-level semantic features for object detection. 

\begin{table}[bpt]\footnotesize
\begin{center}
\begin{tabular}{c|c|c|c|ccc}
\specialrule{1pt}{1pt}{1pt}
\multirow{2}{*}{Networks} & \multirow{2}{*}{Targets} & \multicolumn{2}{c|}{Depth Error (meters)} & \multirow{2}{*}{AP$_\text{3D}$ / AP$_\text{BEV}$ / AP$_{\text{2D}}$} \\ \cline{3-4}
 & & Mean & Median & \\ \hline
% PSMNet* \cite{psmnet} & \multirow{3}{*}{Depth} & & & --- \\
PSMNet-PSV* & \multirow{2}{*}{Depth} & 0.5337 & 0.1093 & ---- \\ 
DSGN &  & \textbf{0.5279} & \textbf{0.1055} & ---- \\ \hline
PSMNet-PSV* & \multirow{2}{*}{Both} & 0.5606 & 0.1157 & 46.41 / 57.57 / 80.67 \\ 
DSGN &  & 0.5586 & 0.1104 & \textbf{54.27} / \textbf{63.91} / \textbf{83.59} \\ 
\specialrule{1pt}{1pt}{1pt}
\end{tabular}
\end{center}
\caption{Influence on depth estimation, evaluated on KITTI val images. PSMNet-PSV* is a variant of PSMNet \cite{psmnet}, which uses one 3D hourglass module instead of three of them for refinement considering limited memory space and takes the plane-sweep approach to construct cost volume.} 
%\vspace{-0.1in}
\label{table: stereo matching}
\end{table}

\section{Conclusion}
We have presented a new 3D object detector on binocular images. It shows that an end-to-end stereo-based 3D object detection is feasible and effective. Our unified network encodes 3D geometry via transforming the plane-sweep volume to a 3D geometric one. Thus, it is able to learn high-quality geometric structure features for 3D objects on the 3D volume. The joint training lets the network learn both \textit{pixel-} and \textit{high-level} features for the important tasks of stereo correspondence and 3D object detection.  

Without bells and whistles, our one-stage approach outperforms other image-based approaches and even achieves comparable performance with a few LiDAR-based approaches on 3D object detection. The ablation study investigates several key components for training 3D volume in Table \ref{table: constraint}. Although the improvement is clear and explained, our understanding of how the 3D volume transformation works will be further explored in our future work.

\section*{A.~~ Appendix}

\subsection*{A.1.~~ More Experiments}

% \subsection*{A.1. More 3D Hourglass on Cost Volume or 3D Geometry Volume?}
% Balance ! To ensure the computation, we use two hourglass network of the same parameters.

\paragraph{Correlation between stereo (depth) and 3D object detection accuracy.} 
Following KITTI stereo metric, we consider a pixel as being correctly estimated if its depth error is less than an \textit{outlier\_thresh}. The percentage of non-outlier pixels inside the object box is deemed as depth estimation precision. 

With several \textit{outlier\_threshes} (0.1, 0.3, 0.5, 1.0, 2.0 meters), scatter plots of stereo and detection precision are drawn. We observed that when \textit{outlier\_thresh} is 0.3 meter, the strongest linear correlation is yielded.

\begin{table}[h]\footnotesize
\begin{center}
\begin{tabular}{c|cccccc}
\hline
\textit{Outlier\_thresh} & $>2$m & $>1$m & $>0.5$m & $>0.3$m & $>0.1$m \\ \hline
PCC & 0.249 & 0.353 & 0.438 & 0.450 & 0.417 \\ \hline
\end{tabular}
\end{center}
\caption{Pearson's correlation coefficients (PCC) for a set of \textit{outlier\_threshes} between depth estimation precision and detection accuracy. \textit{Outlier\_thresh}$=0.3m$ yields the strongest linear correlation.} 
\label{table: pearson}
\end{table}

\begin{figure}[h]
  \begin{center}
    \includegraphics[width=85mm]{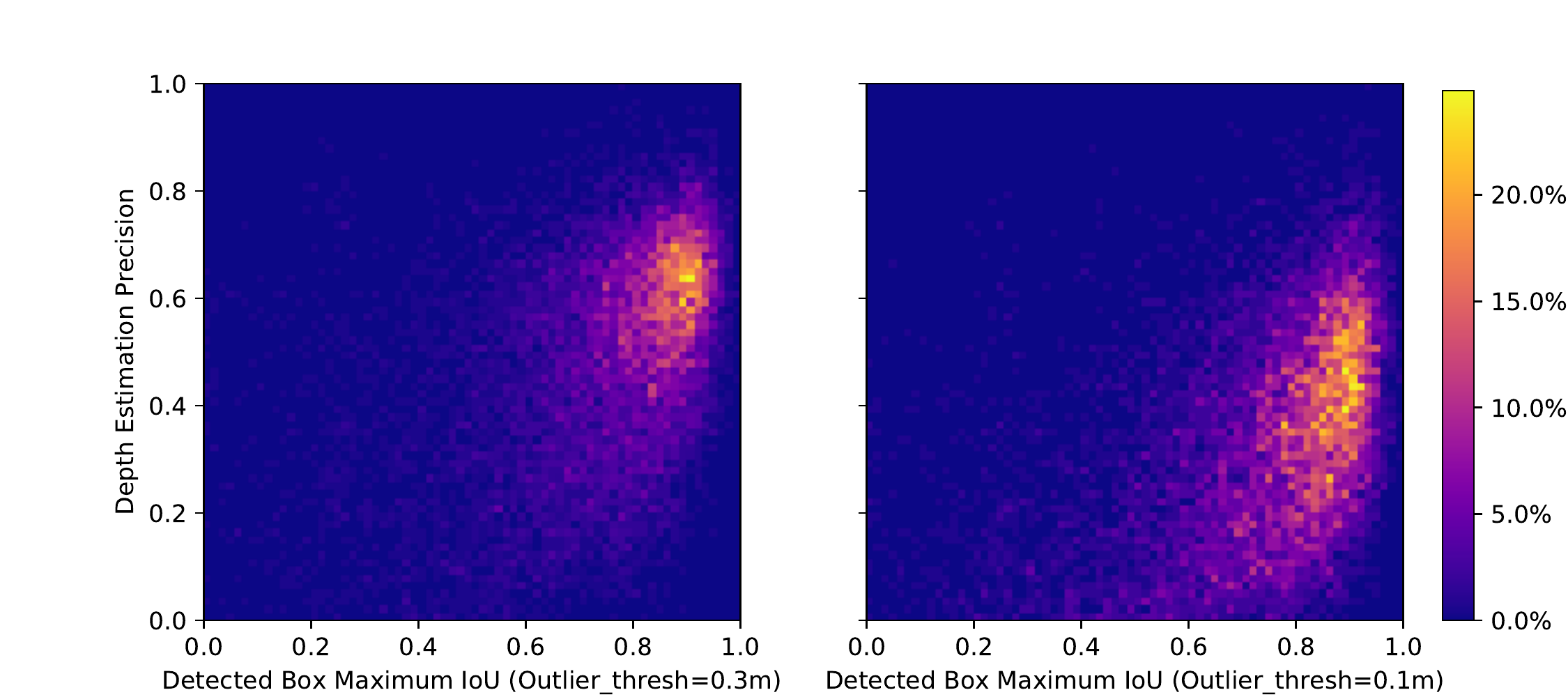}
  \end{center}
  \caption{BEV Detection precision versus depth precision for all predicted boxes for \textsl{Car} on KITTI \textit{val} set. Only TPs with IoU$>0.01$ and score $>0.1$ are shown.}
  \label{fig:stereo_ap}
\end{figure}

% We plot the depth precision and its maximum BEV IoU with ground-truth for all positive predictions. 
Figure \ref{fig:stereo_ap} shows the scatter plots with the \textit{outlier\_thresh} $0.3$m and $0.1$m. Quite a few predictions get over $0.7$ detected precision within a certain range of depth estimation error. This reveals that the end-to-end network gives rise to its ability to detect 3D objects even with a larger depth estimation error. The following 3D detector enables compensation for the stereo depth estimation error by 3D location regression with back-propagation.

\begin{figure}[h]
  \begin{center}
    \includegraphics[width=65mm]{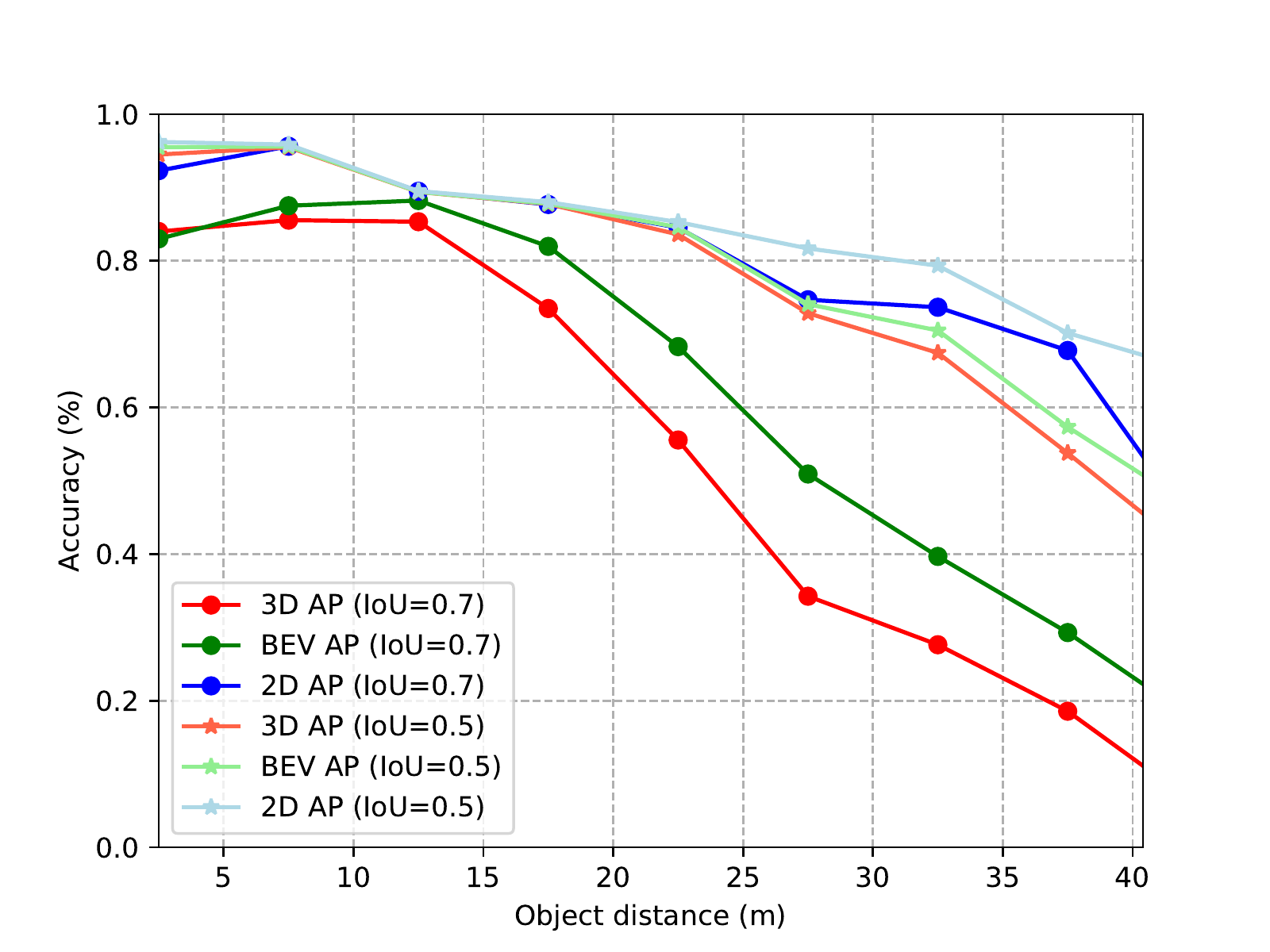}
  \end{center}
  \caption{Detection accuracy versus distance. We separate the range [0, 40] (meters) into 8 intervals, each with 5 meters. All evaluations are conducted within each interval. }
  \label{fig:dist_ap}
\end{figure}

\paragraph{Relationship between distance and detection accuracy.}
Figure \ref{fig:dist_ap} illustrates, as distance increases, all detection accuracy indicators have a shrinking tendency. The average accuracy maintains 80\%+ within 25 meters. In all indicators, 3D AP decreases the fastest, followed by BEV AP and last 2D AP. This observation suggests that the 3D detection accuracy is determined by the BEV location precision beyond 20 meters.

\paragraph{Influence of different features for 3D geometry.}

% Most current depth prediction approaches apply the strategy of ``depth classification'' instead of ``depth regression''. Typical stereo matching networks \cite{gcnet,psmnet,groupwisestereo} intermediately create a cost volume to compute pixel correspondence. 

We discuss the efficiency of different geometric representations for volumetric structure. 
%The meaning of this discussion originates from current depth prediction approaches. 
Most depth prediction approaches \cite{dorn,gcnet,psmnet,groupwisestereo} apply the strategy of ``depth classification'' (such as cost volume) instead of ``depth regression''. Thus, we have several choices for encoding the depth information of cost volume into a 3D volume. The intuitive one is to use a 3D voxel occupancy (denoted by ``Occupancy''). An advanced version is by keeping the probability of voxel occupancy (denoted by ``Probability''). They both have explicit meaning for 3D geometry and can be easily visualized. Another one is by using the last feature map for cost volume as geometric embedding for 3D volume (denoted by ``Last Features'').
%the influence brought by deducting the 3D structure to a concrete  3D geometric data such as a voxel occupancy grid.

\begin{table}[hpt]\footnotesize
\begin{center}
\begin{tabular}{c|cccccc}
\specialrule{1pt}{1pt}{1pt}
Voxel Features & AP$_\text{3D}$ / AP$_\text{BEV}$ / AP$_{\text{2D}}$ \\ \hline
Occupancy & 37.86 / 50.64 / 70.79 \\ 
Probability & 43.24 / 54.87 / 74.93 \\ 
Last Features & \textbf{54.27} / \textbf{63.91} / \textbf{83.59} \\ 
\specialrule{1pt}{1pt}{1pt}
\end{tabular}
\end{center}
\caption{Ablation study on 3D geometric representation. ``Occupancy'' indicates only using binary feature for 3D volume. It is 1 for voxel of minimum cost along the projection ray and is 0 otherwise. ``Probability'' denotes keeping the probability of voxel occupancy instead of quantizing it to 0 or 1. ``Last Features'' represents transforming the last features of cost volume to 3D volume.}
\label{table: depth information loss}
\end{table}

Table \ref{table: depth information loss} reveals the performance gap between ``Occupancy'' / ``Probability''. ``Last Features'' indicates the latent feature embedding (64D) that enables the network to extract more 3D latent geometric information and even semantic cues than the explicit voxel occupancy. It aids learning of 3D structure.

\vspace{-0.05in}
\paragraph{Technical details}

\begin{table}[bpt]\footnotesize
\begin{center}
\begin{tabular}{cccccccc}
\specialrule{1pt}{1pt}{1pt}
JOINT & IMG & ATT & Depth & HG & Flip & AP$_{\text{3D}}$/AP$_{\text{BEV}}$/AP$_\text{2D}$ \\ \hline
 & & & & & & 40.71 / 53.71 / 76.11 \\
\checkmark & & & & & & 45.51 / 56.65 / 78.31 \\
\checkmark & \checkmark & & & & & 44.79 / 56.24 / 81.58 \\
\checkmark & \checkmark & \checkmark & & & & 46.52 / 57.44 / 82.41 \\ 
\checkmark & & & \checkmark & & & 45.79 / 56.89 / 78.49 \\ 
\checkmark & \checkmark & \checkmark & & \checkmark & & 51.73 / 61.74 / 83.6 \\ 
\checkmark & \checkmark & \checkmark & & \checkmark & \checkmark & 54.27 / 63.91 / 83.59 \\ 
\specialrule{1pt}{1pt}{1pt}
\end{tabular}
\end{center}
\caption{Ablation study evaluated in moderate level. ``JOINT'' indicates using joint optimization instead of separable optimization for bounding boxes regression. ``IMG'' denotes concatenating the mapped left image feature to 3DGV for retrieving more 2D semantics. ``ATT (Attention)'' represents concatenating the mapped image feature weighted by the corresponding depth probability. ``Depth'' indicates warping the final matching cost volume to 3DGV. ``HG (Hourglass)'' represents applying an hourglass module in 3DGV. ``Flip'' means using random horizontal flipping augmentation.}
\label{table: technical details}
\end{table}

We explore several technical details used in DSGN and discuss their importance in the pipeline in Table \ref{table: technical details}. Joint optimization of bounding box regression improves accuracy ($+4.80$ AP) than the separable optimization. The intermediate 3D volume representation enables the network to naturally retrieve image feature for more 2D semantic cues. However, 3DGV cannot directly benefit from the concatenation of mapped 32D image features and warped predicted cost volume. Instead, the image feature weighted by the depth probability achieves $+1.01$ AP gain. Further, Involving more computation by an extra hourglass module on 3D object detector and flip augmentation, DSGN finally achieves $54.27$ AP on 3D object detection. 

\begin{table*}[bpt]\footnotesize
\begin{center}
\begin{tabular}{c|c|c|c|c|c|c|c|c|c|c}
\specialrule{1pt}{1pt}{1pt}
\multirow{2}{*}{Modality} & \multirow{2}{*}{Method} & \multicolumn{3}{c|}{ \textbf{3D Detection AP} (\textbf{\%})} & \multicolumn{3}{c|}{BEV Detection AP (\%)} & \multicolumn{3}{c}{2D Detection AP (\%)}\\ \cline{3-11}
& & Easy & \textbf{Moderate} & Hard & Easy & \textbf{Moderate} & Hard & Easy & \textbf{Moderate} & Hard \\ \hline

\multicolumn{11}{c}{\textsl{Pedestrian}} \\ \hline
Mono & M3D-RPN \cite{brazil2019m3d} & -- & 11.09 & -- & -- & 11.53 & -- & -- & -- & -- \\ \hline
\multirow{2}{*}{Stereo} & PL: F-PointNet* \cite{pseudolidar} & 33.8 & 27.4 & 24.0 & 41.3 & 34.9 & 30.1 & -- & -- & -- \\ 
& DSGN (ours) & {\bf 40.16} & {\bf 33.85} & {\bf 29.43} & {\bf 47.92} & {\bf 41.15} & {\bf 36.08} & {\bf 59.06} & {\bf 54.00} & {\bf 49.65} \\ \hline
\multicolumn{11}{c}{\textsl{Cyclist}} \\ \hline

Mono & M3D-RPN \cite{brazil2019m3d} & -- & 2.81 & -- & -- & 3.61 & -- & -- & -- & -- \\ \hline
\multirow{2}{*}{Stereo} & PL: F-PointNet* \cite{pseudolidar} & {\bf 41.3} & {\bf 25.2} & {\bf 24.9} & {\bf 47.6} & {\bf 29.9} & {\bf 27.0} & -- & -- & -- \\ 
& DSGN (ours) & 37.87 & 24.27 & 23.15 & 41.86 & 25.98 & 24.87 & 49.38 & 33.97 & 32.40 \\ 
\specialrule{1pt}{1pt}{1pt}
\end{tabular}
\end{center}
\caption{Comparison of results for \textsl{Pedestrian} and \textsl{Cyclist} on KITTI \textit{val} set. PL: F-PointNet* uses extra Scene Flow dataset to pretrain the stereo matching network.}
\label{table: Pedestrian and Cyclist results on KITTI val set.}
\end{table*}

\begin{table*}[bpt]\footnotesize
\begin{center}
\begin{tabular}{c|c|c|c|c|c|c|c|c|c|c}
\specialrule{1pt}{1pt}{1pt}
\multirow{2}{*}{Modality} & \multirow{2}{*}{Method} & \multicolumn{3}{c|}{ \textbf{3D Detection AP} (\textbf{\%})} & \multicolumn{3}{c|}{BEV Detection AP (\%)} & \multicolumn{3}{c}{2D Detection AP (\%)}\\ \cline{3-11}
& & Easy & \textbf{Moderate} & Hard & Easy & \textbf{Moderate} & Hard & Easy & \textbf{Moderate} & Hard \\ \hline

\multicolumn{11}{c}{\textsl{Pedestrian}} \\ \hline
% PL: F-PointNet \cite{pseudolidar} & 33.8 / 41.3  & 27.4 / 34.9 & 24.0 / 30.1 \\ 
Mono & M3D-RPN \cite{brazil2019m3d} & 4.92 & 3.48 & 2.94 & 5.56 & 4.05 & 3.29 & 56.64 & 41.46 & 37.31 \\ \hline
\multirow{2}{*}{Stereo} & RT3DStereo \cite{rt3dstereo} & 3.28 & 2.45 & 2.35 & 4.72 & 3.65 & 3.00 & 41.12 & 29.30 & 25.25 \\ 
& DSGN (ours) & {\bf 20.53} & {\bf 15.55} & {\bf 14.15} & {\bf 26.61} & {\bf 20.75} & {\bf 18.86} & {\bf 49.28} & {\bf 39.93} & {\bf 38.13} \\ \hline
\multicolumn{11}{c}{\textsl{Cyclist}} \\ \hline
Mono & M3D-RPN \cite{brazil2019m3d} & 0.94 & 0.65 & 0.47 & 1.25 & 0.81 & 0.78 & 61.54 & 41.54 & 35.23 \\ \hline
\multirow{2}{*}{Stereo} & RT3DStereo \cite{rt3dstereo} & 5.29 & 3.37 & 2.57 & 7.03 & 4.10 & 3.88 & 19.58 & 12.96 & 11.47 \\
% PL: F-PointNet \cite{pseudolidar} & -- / -- / -- & -- / & {\bf 24.9} / {\bf 27.0} \\ 
& DSGN (ours) & 27.76 & 18.17 & 16.21 & 31.23 & 21.04 & 18.93 & 49.10 & 35.15 & 31.41 \\ 
\specialrule{1pt}{1pt}{1pt}
\end{tabular}
\end{center}
\caption{Comparison of results for \textsl{Pedestrian} and \textsl{Cyclist} on KITTI test set (official KITTI Leaderboard). }
\label{table: Pedestrian and Cyclist results on KITTI test set.}
\end{table*}

% \subsubsection*{A.1.2 Results on other categories}

\paragraph{\textsl{Pedestrian} and \textsl{Cyclist} detection.}

The main challenges for detecting \textsl{Pedestrian} and \textsl{Cyclist} are the limited data (about only 1/3 of images are annotated) and the difficulty to estimate their poses in an image even for human. As a result, most image-based approaches yield poor performance or are not validated on \textsl{Pedestrian} and \textsl{Cyclist}. Since the evaluation metric is changed on the official KITTI leaderboard, We only report the available results from original papers and the KITTI leaderboard. 
% we do not find the submission item of PL/PL++ \cite{pseudolidar,pseudo++} on the KITTI leaderboard.

Experimental results in Table \ref{table: Pedestrian and Cyclist results on KITTI val set.} shows that our approach achieves better results on \textsl{Pedestrian} but worse ones on \textsl{Cyclist} compared with PL: F-PointNet. We note that PL: F-PointNet used Scene Flow dataset \cite{dispnet} to pre-train the stereo matching network. Besides, PL: F-PointNet achieves the best result on \textsl{Pedestrian} and the model PL: AVOD works best on \textsl{Car} and \textsl{Cyclist}. Table \ref{table: Pedestrian and Cyclist results on KITTI test set.} shows the submitted results on the official KITTI leaderboard.
% Recently LVIS \cite{LVIS} suggests that performance can be biased and of large variance due to the few positive examples in the dataset. 

\subsection*{A.2.~ More Implementation Details}

\paragraph{Network Architecture.}

We show the full network architecture in Table \ref{table:Network Architecture}, including the networks for 2D feature extraction, constructing plane-sweep volume and 3D geometric volume, stereo matching and 3D object detection.

\paragraph{Implementation Details of 3D Object Detector.}

Given the feature map $\mathcal{F}$ on bird's eye view, we put four anchors of different orientation angles $(0, \pi/2, \pi, 3\pi/2)$ on all locations of $\mathcal{F}$. The box sizes of pre-defined anchors used for respectively \textsl{Car}, \textsl{Pedestrian}, \textsl{Cyclist} are $(h_\mathbf{A}=1.56, w_\mathbf{A}=1.6, l_\mathbf{A}=3.9)$, $(h_\mathbf{A}=1.73, w_\mathbf{A}=0.6, l_\mathbf{A}=0.8)$, and $(h_\mathbf{A}=1.73, w_\mathbf{A}=0.6, l_\mathbf{A}=1.76)$. 

The horizontal coordinate $(x_\mathbf{A}, z_\mathbf{A})$ of each anchor lies on the center of each grid in bird's eye view and its center along the vertical direction locates on $y_\mathbf{A} = 0.825$ for \textsl{Car} and $y_\mathbf{A} = 0.74$ for \textsl{Pedestrian} and \textsl{Cyclist}. We set $\gamma = 1$ for \textit{Car} and $\gamma = 5$ for \textsl{Pedestrian} and \textsl{Cyclist} for balancing the positive and negative samples. The classification head of 3D object detector is initialized following RetinaNet \cite{focalloss}. NMS with IoU threshold 0.6 is applied to filter out the predicted boxes on bird's eye view. 

\paragraph{Implementation Details of Differentiable Warping from PSV to 3DGV.}

Let $\mathcal{U} \in \mathbb{R}^{H_I\times W_I\times D_I \times C}$ be the last feature map of PSV, where $C$ is the channel size of features. We first construct a pre-defined 3D volume $\in \mathbb{R}^{H_V \times W_V \times D_V \times 3}$ to store the center coordinate $(x,y,z)$ of each voxel in 3D space (Section \ref{section:implementation}). Then we get the projected pixel coordinate $(u,v)$ by multiplying the projection matrix. $z$ is directly concatenated to pixel coordinate to get $(u,v,z)$ in \textit{camera frustum space}. 

As a result, we get a coordinate volume $\in \mathbb{R}^{H_V \times W_V \times D_V\times 3}$, which stores the mapped coordinates in \textit{camera frustum space}. By \textit{trilinear interpolation}, we fetch the corresponding feature of $\mathcal{U}$ at the projected coordinates to construct the 3D volume $\mathcal{V} \in \mathbb{R}^{H_V \times W_V \times D_V\times C}$, \ie, 3D geometric volume. We ignore the projected coordinates outside the image by setting these voxel features to $0$. In backward operations, the gradient is passed and computed using the same coordinate volume.

\subsection*{A.3.~ Future Work} 

More further studies on stereo-based 3D object detection are recommended here.

\noindent \textbf{The Gap with state-of-the-art LiDAR-based approaches.~~} Although our approach achieves comparable performance with some LiDAR-based approaches on 3D object detection, there remains a large gap with state-of-the-art LiDAR-based approaches \cite{std,pointrcnn,fastpointrcnn,second}. Besides, an obvious problem is the accuracy gap on bird's eye view (BEV) detection. As shown in the table of main results, there is almost 12 AP gap on the moderate and hard level in BEV detection. 

% Some of the difference comes from the annotation because it is annotated according to the object visibility in an image but LiDAR, which can be seen by the LiDAR sensor in a higher location (over 8cm+). But 

% It remains a hard challenge for DSGN to handle highly occluded, truncated and far objects. 
%It remains a hard challenge to increase the robustness of this approach and extract discriminative features for objects. 
One possible solution is high-resolution stereo matching \cite{hsm}, which can help obtain more accurate depth information to increase the robustness for highly occluded, truncated and far objects. 

\noindent \textbf{3D Volume Construction.~~} Table \ref{table: constraint} shows basic comparison of volume construction in DSGN. We expect a more in-depth analysis of the volume construction from multi-view or binocular images, which serves as an essential component design for 3D object understanding. Besides, the effectiveness of 3D volume construction methods still requires more investigation since it needs to balance and provide both depth information and semantic information.

\noindent \textbf{Computation Bottleneck.~~} The computation bottleneck of DSGN locates on the computation of 3D convolutions for computing cost volume. Recent stereo matching work \cite{hierarchical,anytimestereo} focused on accelerating the computation of cost volume. Another significant aspect of constructing cost volume is that current cost volume \cite{gcnet, psmnet} is designed for regressing disparity but not depth. Further research might explore more efficient feature encoding for the plane-sweep cost volume.

\noindent \textbf{Network Architecture Design.~~} There is a trade-off between stereo matching network and 3D detection network for balancing the feature extraction of pixel- and high-level features, which can be conducted by recent popular Network Architecture Search (NAS).

\noindent \textbf{Application on Low-speed Scenario.~~} Our approach shows comparable performance with the LiDAR-based approach on 3D and BEV detection in the close range in the KITTI easy set. Most importantly, it is affordable even with one strong GPU Tesla V100 (\$11,458 (USD)) compared with the price of a 64-beam LiDAR \$75,000 \cite{pseudo++}. It is a promising application of image-based autonomous driving system for low-speed scenarios.

\subsection*{A.4.~ Qualitative Results}

We provide a video demo \footnote{\url{https://www.youtube.com/watch?v=u6mQW89wBbo}} for visualization of our approach, which shows both the detected 3D boxes on front view and bird's eye view. The ground-truth LiDAR point cloud is shown on bird's eye view. The detection results are obtained by DSGN trained on KITTI training split only. The unit of the depth map is meter. 

Some noise observed in the predicted depth map is mainly caused by the implementation details. (1) Noise in the near and far part: 3D volumes are constructed in [2, 40.4] (meters). (2) Noise and large white zone in the higher region ($>$3m): The stereo branch is trained with a sparse GT depth map (64 lines around [-1,3 (meters) along the gravitational $z$-axis, captured by a 64-ray LiDAR).

\begin{table}[hpt]
    \begin{center}
        \includegraphics[width=73mm]{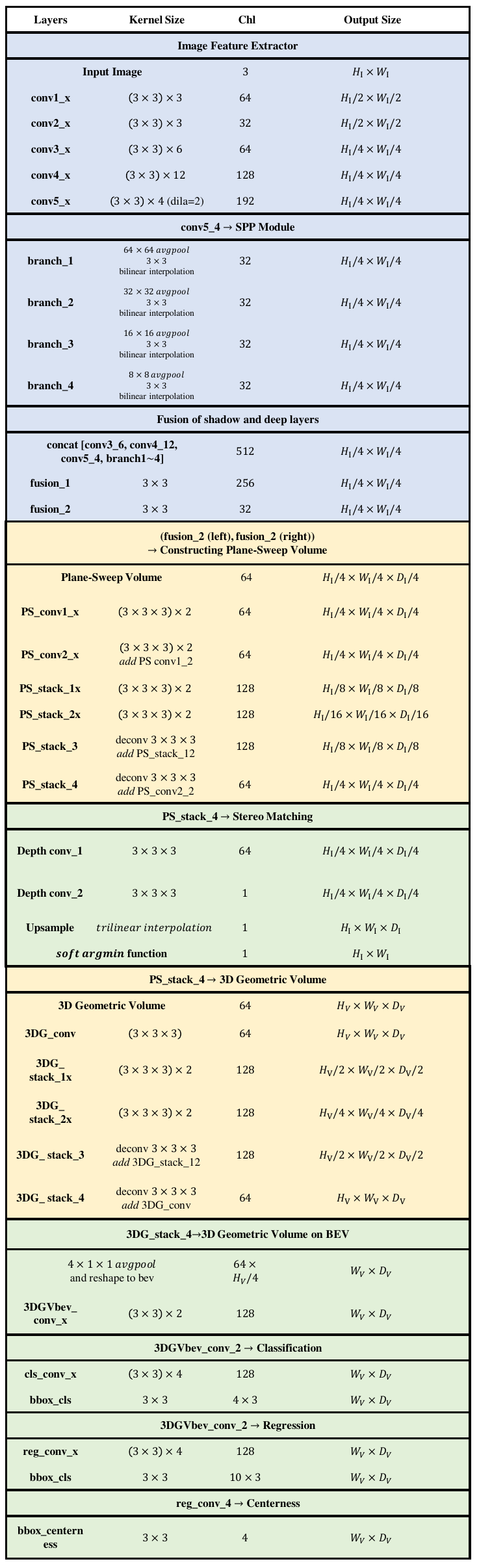}
    \end{center}
    \caption{Full network architecture of DSGN. The color of the table highlights different components.}
    \label{table:Network Architecture}
\end{table}

{\small
\bibliographystyle{ieee_fullname}
\bibliography{egbib}
}
\clearpage

\end{document}